\documentclass{article} 
\usepackage{iclr2027_conference,times}

\usepackage{amsmath,amsfonts,bm}

\def\eqref#1{equation~\ref{#1}}

\def\1{\bm{1}}

\DeclareMathAlphabet{\mathsfit}{\encodingdefault}{\sfdefault}{m}{sl}
\SetMathAlphabet{\mathsfit}{bold}{\encodingdefault}{\sfdefault}{bx}{n}

\usepackage{hyperref}
\usepackage{url}
\usepackage{graphicx}

\usepackage{booktabs}
\usepackage{graphicx}
\usepackage{makecell}
\usepackage{caption}
\usepackage{subcaption}

\usepackage{tikz}
\usepackage{graphicx}
\usepackage{wrapfig}
\usepackage{needspace}
\usepackage{booktabs}
\usepackage{makecell}
\usepackage{graphicx}
\usepackage{enumitem}

\usepackage{amsthm}

\usepackage{algorithm}
\usepackage{algpseudocode}
\usepackage{amsmath}
\usepackage{multirow}

\usetikzlibrary{arrows.meta,positioning,calc}

\definecolor{NatureBlue}{HTML}{3C5488}
\definecolor{NatureGreen}{HTML}{00A087}
\definecolor{NatureOrange}{HTML}{E64B35}
\definecolor{NaturePurple}{HTML}{8491B4}
\definecolor{NatureGray}{HTML}{4A4A4A}

\title{AHEAD: Adaptive Neural Acceleration for Large-Scale Linear Programming}

\author{%
{\bfseries Sihao Liu$^{1}$, Zhichao Hou$^{1}$,
Huayue Gu$^{2}$, Raiyan Rahman$^{1}$,}\\
{\bfseries Liang Wu$^{3}$, Ruozhou Yu$^{1}$,
Xiaorui Liu$^{1}$}\\
$^{1}$North Carolina State University
{\small\texttt{\{sliu78,zhou4,rrahman,ryu5,xliu96\}@ncsu.edu}}\\
$^{2}$Kennesaw State University
{\small\texttt{hgu2@kennesaw.edu}}\\
$^{3}$Nokia Applied Research
{\small\texttt{liang.wu@nokia.com}}
}

\newcommand{\refeq}[1]{Equation~(\ref{#1})}

\usepackage{amsthm}
\usepackage{subcaption}

\iclrfinalcopy 
\begin{document}

\maketitle

\lhead{}

\begin{abstract}
Large-scale linear programs with millions of variables and constraints strain time and memory budgets. First-order methods such as the primal-dual hybrid gradient scale to this regime because they avoid matrix factorizations, but they require many iterations. Learned accelerators can reduce these iterations, yet fewer iterations do not guarantee a shorter solve. Learning inference also takes time, and running the accelerator longer trades solver time for inference time. Existing unrolled accelerators handle this trade-off poorly: each layer processes every variable and constraint, the predicted iterates still need solver refinement, and every instance runs the same layers. We therefore ask: What neural accelerator design achieves end-to-end speedups on large-scale linear programs across sizes and target accuracies? We introduce AHEAD, an accelerator that predicts step sizes for first-order linear programming solvers while keeping their update rules. Its weight-shared predictor reads compact solver statistics and rescales the primal and dual step sizes, so each warm-start iteration stays inexpensive and a short warm-start phase hands the solver a warm start. A learned handoff policy sets the number of warm-start iterations for each instance and target accuracy, after which the solver finishes under its own termination criterion. On packing linear programs with up to $10^7$ variables, AHEAD achieves $6.5$--$11.6\times$ end-to-end speedups over Gurobi's first-order solver from a cold start, including on instances of unseen sizes and distributions. The gains extend to PageRank linear programs and hold across difficulty levels and target accuracies. 
\end{abstract}
\section{Introduction}
\label{sec:intro}
Linear programming (LP) underpins large-scale decision-making, but instances with millions of variables and constraints strain time and memory budgets~\citep{dantzig1963linear,nesterov2014subgradient}. First-order methods (FOMs) are attractive at this scale because they rely on sparse matrix-vector multiplications rather than costly factorizations~\citep{lin2021admm,applegate2023faster,lu2025cupdlp}. However, they often need many iterations, particularly at high accuracy~\citep{applegate2023faster}. Their progress depends strongly on primal and dual step sizes: conservative choices slow convergence, while aggressive ones can destabilize it~\citep{goldstein2013adaptive,malitsky2018first}.

Learning-to-optimize (L2O) offers a promising way to reduce the computational cost of iterative optimization. Early learned optimizers replace hand-designed update rules with learned dynamics~\citep{li2016learning,andrychowicz2016learning,wichrowska2017learned}. More recent solver-integrated methods predict one-shot warm starts~\citep{sambharya2023learning, gao2024certified}, unroll first-order or interior-point trajectories~\citep{li2024pdhg,qian2024exploring, hou2024robust,hou2024protransformer,hou2025robustness}, or learn feedback policies that tune algorithmic parameters online~\citep{oshin2026deep}. Across these approaches, learning exploits recurring structure among related instances to reduce the cost subsequently performed by an optimization solver.

Fewer solver iterations, however, do not guarantee a shorter solve. In learned acceleration of LP solvers, a solve typically has two stages: a neural accelerator first produces a warm start, and the classical solver then continues from it until the solution reaches the target accuracy. We call the two stages together, the neural warm start followed by the classical solve, \emph{end-to-end neural acceleration}. Existing unrolled accelerators typically rely on a fixed number of computational layers. Each learned layer processes every variable and constraint and predicts new primal--dual iterates in place of the solver's native update, causing inference cost to grow with both instance size and rollout depth~\citep{qian2024exploring,li2024pdhg}. These approximate outputs may still require additional solver running time to recover feasibility or attain high accuracy~\citep{sambharya2023learning,gao2024ipmlstm}. Moreover, fixed-depth models use the same compute budget for every instance. This prevents them from adapting computational efforts to instance-specific needs and limits generalization across problem distributions and difficulty levels~\citep{chen2022learning,sucker2025generalization}. Consequently, neural inference and subsequent refinement can offset or even outweigh the time saved by reducing solver iterations. These limitations lead to our central question:
\textbf{What neural accelerator design achieves end-to-end speedups on large-scale LPs across sizes and target accuracies?}

To address this question, we introduce AHEAD (\textbf{A}daptive \textbf{H}orizon for \textbf{E}nd-to-end \textbf{A}cceleration of Primal--\textbf{D}ual Solvers), a lightweight neural accelerator for first-order LP solvers. In the first stage, which we call the warm-start phase, AHEAD applies a weight-shared
step-size predictor at every iteration and accelerates the solver only by
adjusting its step sizes. To keep each warm-start iteration cheap, the predictor
reads only ten statistics built from the solver's own intermediate results and outputs one multiplier for each of the primal and dual step sizes. 
AHEAD's solution quickly approaches the target accuracy on its own. Because only the step sizes change, AHEAD is also fully aligned with the solver: when high accuracy is required, its solution can be handed directly to the solver as a
warm start. A learned handoff policy picks this moment by predicting whether stopping now or continuing gives a shorter end-to-end time. Together, these designs address three core challenges: they let the solver reach the target accuracy in fewer iterations, keep the time spent on each iteration low, and adapt the
number of warm-start iterations to each instance and accuracy.

\begin{figure}
    \centering
    \includegraphics[width=0.95\linewidth]{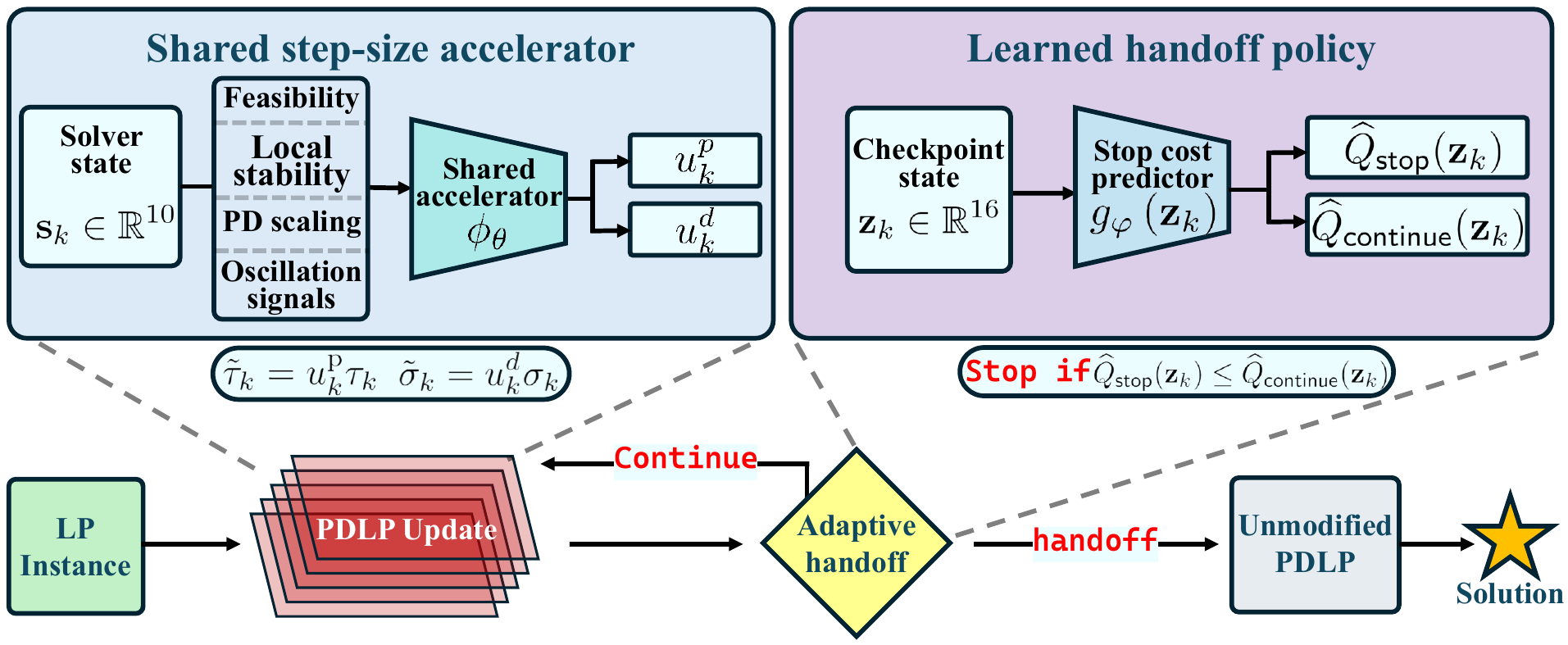}
    \caption{Framework of AHEAD.}
    \label{fig:framework}
\end{figure}

Our main contributions are summarized as follows:
\begin{itemize}[leftmargin=*,labelindent=0pt]
\item \textbf{End-to-end speedups from fast convergence.} AHEAD's
solution quickly approaches the target accuracy within a short warm-start phase, leaving little remaining solver time, which yields many-fold end-to-end speedups across unseen scales, LP families, and accuracy levels.

\item \textbf{Low time per iteration.} We keep the inference time of each warm-start iteration small at any problem size: the predictor reads a 10-dimensional solver-native state and outputs two step-size multipliers, which suffice to steer PDHG through its oscillatory early phase.

\item \textbf{Adaptive number of warm-start iterations.} We apply a weight-shared predictor at each iteration recurrently, so the warm-start phase can run for any number of iterations, and a learned handoff policy selects this number for each instance and accuracy, further reducing the end-to-end time.
\end{itemize}

\section{Related Work}

\textbf{First-order methods for large-scale linear programming.} \citet{chambolle2011first} introduced PDHG, whose matrix--vector updates suit large-scale saddle-point problems. \citet{applegate2021practical} developed PDLP, preconditioning, adaptive step sizes, and primal--dual balancing; subsequent work used adaptive restarting to improve high-accuracy convergence~\citep{applegate2023faster}. These solvers adapt through hand-designed rules, but do not learn from related LP solves.

\textbf{Learning components of LP solvers.} \citet{li2022reformulate} learn variable reorderings, while \citet{kuang2023accelerate} learn presolve routines, including when to stop presolving. \citet{iwata2025learning} generate instance-specific projections to reduce LP dimension. Other works learns simplex bases or pivot decisions~\citep{fan2023smart,liu2024pivot,lu2025learning}. These methods can accelerate stages such as presolve or simplex search, but their neural inference overhead may outweigh the solver time saved.

\textbf{Learned optimizers and warm starts.} Learned update rules can transfer across optimization tasks~\citep{andrychowicz2016learning,wichrowska2017learned}, although generalization outside the training distribution remains challenging~\citep{liu2023structures}. \citet{sambharya2023learning} learn warm starts for quadratic optimization. For LPs, \citet{li2024pdhg} use a fixed-depth PDHG-Net to predict primal--dual states, then refine them with PDLP; they also demonstrate transfer to larger instances of related problem families. \citet{oshin2026deep} learn dimension-agnostic parameter feedback for QP and SQP solvers.

\textbf{Adaptive computation.} \citet{graves2016adaptive,banino2021pondernet, hou2026fine} study input-dependent neural computation, while \citet{chen2020stop} learn when to stop the neural network. For LPs, however, stopping decisions should exploit LP-specific feasibility and gap signals and account for the remaining solver work after learned computation ends. Taken together, prior work leaves open how to combine inexpensive control of LP iterations with instance-dependent computation to obtain consistent end-to-end gains across problem sizes and families.

\section{AHEAD: Lightweight Warm Starts with Adaptive Depth}
\label{sec:method}

For an LP instance $i$, consider a learned accelerator parameterized by $\theta$ that controls the solver during a warm-start phase of $K$ iterations, and then hands the solver a primal--dual warm start $\mathbf{z}_{i,K}=(\hat{\mathbf{x}}_i^K,\hat{\mathbf{y}}_i^K)$; we call this step the \emph{handoff}.
The solver continues from $\mathbf{z}_{i,K}$ until it reaches the target accuracy $\epsilon$.
We define the end-to-end time of instance $i$ is
\begin{equation}
  \mathcal{T}_i(\theta,K;\epsilon)
  =
  T_{\mathrm{warm}}(K)
  +
  T_{\mathrm{solver}}(\mathbf{z}_{i,K};\epsilon),
  \label{eq:total-cost}
\end{equation}
where the warm-start time $T_{\mathrm{warm}}(K)$ covers the $K$ warm-start iterations, including the inference of the predictor and the handoff policy, and the remaining solver time $T_{\mathrm{solver}}(\mathbf{z}_{i,K};\epsilon)$ covers the solver's run from the warm start until it reaches $\epsilon$. The cold-start time $\mathcal{T}_{i,\mathrm{cold}}=\mathcal{T}_i(\theta,0;\epsilon)$ covers the same run from the solver's default initial point.

AHEAD yields an end-to-end speedup on instance $i$ if $\mathcal{T}_i(\theta,K_i;\epsilon <\mathcal{T}_{i,\mathrm{cold}}$, where $K_i$ is the number of warm-start iterations selected for instance $i$. Our goal is to achieve this on all or most instances, so we design AHEAD to keep both terms of \refeq{eq:total-cost} small. Section~\ref{sec:cost} derives what each term requires of the accelerator. Based on this analysis, AHEAD consists of two learned components: a weight-shared step-size predictor $f_\theta$ (Section~\ref{sec:how-far}) that adjusts the primal and dual step sizes during the warm-start phase, and a learned handoff policy $g_\varphi$ (Section~\ref{sec:how-long}) that selects the number of warm-start iterations $K_i$ for each instance and accuracy.
Section~\ref{sec:training} describes how we train the predictor over varying numbers of warm-start iterations, and Figure~\ref{fig:framework} summarizes the framework.

\subsection{Design Requirements}
\label{sec:cost}

Reducing $\mathcal{T}_i$ requires keeping both terms of \refeq{eq:total-cost} small.
To make warm-starting worthwhile, its runtime must stay small: the accelerator runs each warm-start iteration, so its input must be cheap to compute and fixed-dimensional regardless of problem size. We consider LPs in bounded form
\begin{equation}
  \min_{x}\; c^\top x
  \quad \text{s.t.} \quad
  Gx \ge h, \quad Ax = b, \quad \ell \le x \le u,
  \label{eq:lp}
\end{equation}
where the bounds $\ell$ and $u$ may contain infinite entries. PDHG~\citep{chambolle2011first} is a classical first-order method for~\refeq{eq:lp}. Let $\mathcal{K} = [G; A] \in \mathbb{R}^{m \times n}$,
$q = [h; b]$, and $\mathcal{Y} = \mathbb{R}_+^{m_1} \times \mathbb{R}^{m_2}$ for
the $m_1$ inequality and $m_2$ equality rows, with $m = m_1 + m_2$. PDHG alternates
\begin{equation}
    x_{k+1} = \mathrm{proj}_{[\ell,u]}\big (x_k - \tau_k (c - \mathcal{K}^\top y_k)\big), \quad
  y_{k+1} = \mathrm{proj}_{\mathcal{Y}}\big (y_k + \sigma_k (q - \mathcal{K}(2x_{k+1} - x_k)) \big),
\label{eq:pdhg} 
\end{equation}
where $\mathrm{proj}_{\mathcal{C}} (v) := \arg\min_{w \in \mathcal{C}} \frac{1}{2}\|w - v\|_2^2$. 
PDHG-based solvers such as PDLP first go through an oscillatory exploration phase and only then converge toward a solution~\citep{applegate2023faster}. 
This observation motivates a lightweight step-size predictor that adjusts $\tau_k$ and $\sigma_k$ at every iteration to shorten this early phase. After $K$ warm-start iterations, the accelerator hands the warm start to the solver, which continues until it reaches $\epsilon$. The best number of warm-start iterations $K_i^\star = \arg\min_{K} \mathcal{T}_i(\theta,K;\epsilon)$, balances the two terms and depends on both the instance and $\epsilon$. However, $K_i^\star$ is known only in hindsight, since evaluating it requires running solver to termination from the warm start at every candidate $K$. A practical accelerator therefore needs an online rule that selects $K_i$.

Formally, AHEAD keeps \refeq{eq:pdhg} and only rescales its step sizes:
\begin{equation}
  (\tilde{\tau}_k, \tilde{\sigma}_k) =
  \begin{cases}
    (u_k^{\mathrm p}\tau_k,\; u_k^{\mathrm d}\sigma_k), & k < K_i,\\
    (\tau_k,\; \sigma_k), & k \ge K_i,
  \end{cases}
  \label{eq:decomp}
\end{equation}
where $\tau_k, \sigma_k$ are the solver's own step sizes and $u_k^{\mathrm p}, u_k^{\mathrm d} > 0$ are learned step-size multipliers. AHEAD thus makes two online decisions: it predicts $(u_k^{\mathrm p}, u_k^{\mathrm d})$ at every warm-start iteration and selects $K_i$ for each instance and $\epsilon$. The predictor outputs only two scalars per iteration, and the iterate at every warm-start iteration remains a valid PDHG iterate.

\subsection{Weight-Shared Step-Size Predictor}
\label{sec:how-far}
To predict $(u_k^{\mathrm p}, u_k^{\mathrm d})$ at each warm-start iteration, we design a weight-shared step-size predictor (Figure~\ref{fig:framework}, left). It maps the current solver state to the two multipliers and shares its weights across iterations. We first describe the state and then the predictor.

\textbf{Solver state.} At each warm-start iteration $k$, the predictor reads a state $\mathbf{s}_k$ that describes the current progress of PDHG. To choose the step sizes for the next iteration, the state must answer four questions: how much the current iterate violates the primal and dual constraints, whether a larger step would remain stable, how to split the step between primal and dual, and whether recent steps make progress or oscillate. PDLP~\citep{applegate2026pdlp} already answers each question with its own mechanism: its termination residuals, its step-acceptance check, its primal weight, and its restart rule (Appendix~\ref{app:pdlp}). We therefore build one group of features from each mechanism, reusing only the solver's own intermediate results, so that $\mathbf{s}_k$ is cheap to compute and has the same dimension for every LP, as required in Section~\ref{sec:cost}:
\begin{equation}
  \mathbf{s}_k =
  \Big[
    \underbrace{\bar r_k^{\mathrm p}, \bar r_k^{\mathrm d}}_{\text{feasibility}},\;
    \underbrace{\mathcal{M}_k, \mathcal{N}_k, \mathcal{U}_k, \mathcal{S}_k}_{\text{local stability}},\;
    \underbrace{\bar\omega_k}_{\text{primal--dual scaling}},\;
    \underbrace{C_k^x, C_k^y, \rho_k}_{\text{local trajectory}}
  \Big] \in \mathbb{R}^{10}.
  \label{eq:state}
\end{equation}

\textbf{Feasibility.} A primal--dual pair $(x, y)$ of an LP is optimal if and only if $x$ is primal feasible, $y$ is dual feasible, and the duality gap is zero. The first group tracks the two feasibility conditions: the primal residual $\bar r_k^{\mathrm p} = \log\big(1 + \|Ax_k - b\|_2/\sqrt{m}\big)$ measures how much $x_k$ violates the primal constraints, and the dual residual $\bar r_k^{\mathrm d} = \log\big(1 + \|c - A^\top y_k\|_2/\sqrt{n}\big)$ measures how much $y_k$ violates the dual constraints. Large residuals indicate an early phase of the solve, where larger steps pay off, and small residuals call for finer steps. 

\textbf{Local stability.} The second group tells the predictor whether a larger step would remain stable. With constant step size, PDHG converges when $\tau\sigma\|\mathcal{K}\|_2^2 < 1$~\citep[Theorem~1]{chambolle2011first}. This bound is often too conservative in practice. 
PDLP~\citep{applegate2021practical} instead adapts the step size with a local \emph{stability condition} on the actual movement: it computes the update in \refeq{eq:pdhg} with a trial step size, keeps the update if the condition holds, and otherwise shrinks the step size and recomputes the update (Appendix~\ref{app:pdlp}).
Let $\Delta x_k = x_k - x_{k-1}$ and $\Delta y_k = y_k - y_{k-1}$ denote the change between the last two iterations, and write $\tau_k = \eta_k/\omega_k$ and $\sigma_k = \eta_k\omega_k$, where $\eta_k > 0$ sets the overall step size and $\omega_k > 0$ is the solver's primal weight. With the weighted movement $\mathcal{M}_k = \frac{\omega_k}{2}\|\Delta x_k\|_2^2 + \frac{1}{2\omega_k}\|\Delta y_k\|_2^2$
and the primal--dual interaction $\mathcal{N}_k = \langle A\Delta x_k, \Delta y_k\rangle$,
PDLP's condition can be written as
\begin{equation}
  \mathcal{M}_k/\eta_k - \mathcal{N}_k \ge 0 .
  \label{eq:local-check}
\end{equation}
This condition always holds when $\mathcal{N}_k \le 0$. When $\mathcal{N}_k > 0$, it limits the step size to $\eta_k \le \mathcal{M}_k/\mathcal{N}_k$. We define the utilization as the ratio of the current step size to this upper limit, $\mathcal{U}_k = \eta_k / (\mathcal{M}_k/\mathcal{N}_k) = \eta_k\mathcal{N}_k/\mathcal{M}_k$, so the condition holds exactly when $\mathcal{U}_k \le 1$.
We therefore include $\mathcal{M}_k$, $\mathcal{N}_k$, $\mathcal{U}_k$, and the bounded ratio $\mathcal{S}_k = \tanh(\mathcal{N}_k/\mathcal{M}_k)$ in the state: $\mathcal{U}_k$ tells the predictor how close the current step size is to this limit, and the sign of $\mathcal{S}_k$ tells it whether the condition can be violated at all. The predictor thus observes the same condition that the classical solver imposes on each update, and can set the next multipliers.

\textbf{Primal--dual scaling.} The third group tells the predictor how to split the step between primal and dual. The solver sets this split through its primal weight: $\tau_k = \eta_k/\omega_k$ and $\sigma_k = \eta_k\omega_k$. The best pair of multipliers thus depend on $\omega_k$, so we include the detrended log-weight $\bar\omega_k = \log\omega_k - 0.8\log(n/m)$, which treats reciprocal primal and dual scalings symmetrically.

\textbf{Local trajectory.} The last group tells the predictor whether recent steps make progress or oscillate. PDHG iterates often spiral around saddle points before settling, which slows early progress~\citep{chambolle2011first, applegate2023faster}. PDLP counters this with restarts triggered by the normalized duality gap, which measures how far an iterate is from the saddle point but needs an extra optimization problem to compute, so PDLP evaluates it only periodically~\citep{applegate2021practical}. However, our predictor needs this information at every iteration, and computing the gap that often would be too expensive. 
We therefore measure the directional consistency of consecutive updates with cosine similarities, which are inexpensive to compute at every iteration:
\begin{equation}
  C_k^x = \frac{\langle \Delta x_k, \Delta x_{k-1}\rangle}
               {\|\Delta x_k\|_2 \|\Delta x_{k-1}\|_2 + \delta},
  \qquad
  C_k^y = \frac{\langle \Delta y_k, \Delta y_{k-1}\rangle}
               {\|\Delta y_k\|_2 \|\Delta y_{k-1}\|_2 + \delta},
  \label{eq:cosine}
\end{equation}
where the small constant $\delta > 0$ prevents division by zero. Up to $\delta$, each cosine equals $+1$ when consecutive updates point in the same direction and $-1$ when they reverse, and it does not change when either update is rescaled, for example by a new step size. Values near $+1$ thus indicate steady progress and suggest that a larger step is safe, whereas values near $-1$ indicate oscillation and need to adapt the step smaller. We also include the movement ratio $\rho_k = \log(\|\Delta z_k\|_2 / \|\Delta z_{k-1}\|_2)$ with $\Delta z_k = (\Delta x_k, \Delta y_k)$, which shows whether the updates are growing ($\rho_k > 0$) or shrinking ($\rho_k < 0$). These features give the predictor the progress information that PDLP's restart rule relies on. Thus, AHEAD itself never restarts during the warm-start phase.

\textbf{Weight-shared step-size predictor.} Given $\mathbf{s}_k$, the predictor outputs the logarithms of the two step-size multipliers,
\begin{equation}
  \big[\ell_k^{\mathrm p},\, \ell_k^{\mathrm d}\big]^\top = f_\theta(\mathbf{s}_k),
  \qquad
  u_k^{\mathrm p} = e^{\ell_k^{\mathrm p}}, \quad u_k^{\mathrm d} = e^{\ell_k^{\mathrm d}},
  \label{eq:predictor}
\end{equation}
where $f_\theta$ is a small MLP with tanh activations. Predicting logarithms keeps both multipliers positive, and \refeq{eq:decomp} applies them to the solver's native step sizes.

The primal--dual update directions of the solver are faithfully kept in AHEAD, as we find them to be efficient and reliable directions towards optimality.
The primal and dual step sizes are much harder to tune, since first-order methods like PDLP lack the higher-order information as interior-point methods to find the optimal step size for each update.
Hand-designed rules either overestimate or underestimate the optimal step size.
Learning the optimal step sizes only can thus reap the most benefit of learning without introducing more model complexity that lowers per-iteration efficiency.

Another core design principle is a looped architecture for recursive solution refinement: we apply the same predictor $f_\theta$ at every warm-start iteration. This is a paradigm shift from existing neural solvers with a fixed model depth and independent model parameters per-layer: they require more parameters and cannot adapt to the varying number of iterations. AHEAD runs like a classical solver: it recursively refines primal-dual solutions, without a pre-determined number of iterations, until the handoff policy ends the warm-start phase. This adaptive recursion is key to generalizability across instances with different sizes and difficulties, as reported in other literature such as~\citep{chen2020stop} and recurrent networks~\citep{schwarzschild2021learnalgorithmgeneralizingeasy}.

\subsection{Learned Handoff Policy}
\label{sec:how-long}

With the weight-shared predictor, AHEAD can run for any number of warm-start iterations. For moderate accuracy requirements, such as $\epsilon = 10^{-4}$, the warm-start iterations alone already bring the solution close to the accuracy. Because AHEAD is fully aligned with the solver, it enables a second capability: when higher accuracy is required, AHEAD hands its solution directly to the solver as warm start. This warm start substantially reduces the number of iterations the solver needs to converge at high accuracy, while still achieving a large end-to-end speedup over a cold start.

A key question in this handoff is when to stop the warm-start phase and switch to the solver's own iterations. 
Switching too late wastes warm-start iterations that no longer reduce the remaining solver time, whereas switching too early gives the solver a weaker warm start than AHEAD could provide. 
We address this with a learned handoff policy $g_\varphi$ (Figure~\ref{fig:framework}, right). It selects the number of warm start iterations $K_i$ for any given target accuracy $\epsilon$. Since the best choice $K_i^\star$ is known only in hindsight (Section~\ref{sec:cost}), the policy makes this decision online. At every warm-start iteration $k$, it reads a \emph{handoff state} $\mathbf{h}_k$, predicts the end-to-end time of stopping now and of continuing, and ends the warm-start phase once stopping is predicted to be faster. Unlike halting in adaptive neural computation~\citep{graves2016adaptive,banino2021pondernet}, this decision depends on the remaining solver time after handoff, not on the quality of a network output.

\textbf{Handoff state.} Whether to handoff depends on how close the solver is to convergence and how fast it is approaching it. The solver monitors the primal residual, dual residual, and duality gap,
\begin{equation}
  e_k^{\mathrm p} = \frac{\|Ax_k - b\|_2}{1+\|b\|_2}, \quad
  e_k^{\mathrm d} = \frac{\|c - A^\top y_k\|_2}{1+\|c\|_2}, \quad
  e_k^{\mathrm g} = \frac{|c^\top x_k - b^\top y_k|}{1 + |c^\top x_k| + |b^\top y_k|},
  \label{eq:termination}
\end{equation}
and reaches target accuracy $\epsilon$ when
$\max\{e_k^{\mathrm p}, e_k^{\mathrm d}, e_k^{\mathrm g}\} \le \epsilon$~\citep{applegate2021practical}. Convergence is decided by the slowest criterion, and the same residual may be close enough for a loose $\epsilon$ but far from a tight one. We therefore measure, for each criterion, the log-gap to the target accuracy
\begin{equation}
  \Delta_k^j(\epsilon) = \log\frac{e_{k}^j + \delta}{\epsilon + \delta},
  \qquad j \in \{\mathrm p, \mathrm d, \mathrm g\},
  \label{eq:progress}
\end{equation} 
where $\delta > 0$ avoids $\log 0$. Because $\epsilon$ is considered in the handoff state, the same learned handoff policy can adapt the number of warm-start iterations across target accuracies.

\textbf{Handoff rule.} At warm-start iteration $k$, AHEAD can either hand off now or continue. An MLP $g_\varphi$ maps the handoff state $\mathbf{h}_k$ to
\begin{equation}
  \big[\widehat{Q}_{\mathrm{handoff}}(\mathbf{h}_k),\,
       \widehat{Q}_{\mathrm{continue}}(\mathbf{h}_k)\big] = g_\varphi(\mathbf{h}_k),
  \label{eq:qhead}
\end{equation}
where $\widehat{Q}_{\mathrm{handoff}}$ and $\widehat{Q}_{\mathrm{continue}}$ estimate the end-to-end time of handing off now and of handing off at the best later iteration, respectively. Both are divided by the cold-start time, so they are comparable across instances of different sizes and difficulty. 
AHEAD hands off at the first warm-start iteration $k$ with $
     \widehat{Q}_{\mathrm{handoff}}(\mathbf{h}_k) \le \widehat{Q}_{\mathrm{continue}}(\mathbf{h}_k),$ or after $K_{\max}$ iterations if this condition never holds.
estimate the same quantity, so the rule needs no extra threshold. After the handoff, the solver continues from the warm start with its own step-size rule and termination criterion, so every returned solution meets the target accuracy $\epsilon$, exactly as a cold-start solve does.

\subsection{Multi-Horizon Training}
\label{sec:training}

To serve different accuracy requirements, we train the predictor over varying horizons $K$ rather than a fixed one. The handoff policy may end learned acceleration at any checkpoint, so the predictor must leave a useful iterate at every candidate horizon. Formally, let $\mathcal{H}$ be the set of candidate horizons and $(\mathbf{x}_i^\star, \mathbf{y}_i^\star)$ an optimal primal--dual solution of instance $i$. We train the predictor by
\begin{equation}
  \min_\theta\;
  \mathbb{E}_{i\sim\mathcal{D}}\,
  \mathbb{E}_{K\sim\mathcal{U}(\mathcal{H})}
  \Bigg[
    \frac{\|\hat{\mathbf{x}}_i^K - \mathbf{x}_i^\star\|_2^2}{\|\mathbf{x}_i^\star\|_2^2 + \delta}
    +
    \frac{\|\hat{\mathbf{y}}_i^K - \mathbf{y}_i^\star\|_2^2}{\|\mathbf{y}_i^\star\|_2^2 + \delta}
  \Bigg],
  \label{eq:multi-horizon}
\end{equation}
where $(\hat{\mathbf{x}}_i^K, \hat{\mathbf{y}}_i^K)$ is the handoff iterate after $K$ learned iterations. The normalized distance to the optimum serves as a differentiable proxy for the remaining solver work $T_{\mathrm{solver}}$. This objective can be viewed as training over a distribution of horizons rather than a single fixed depth. By sampling $K$, the predictor learns a state-dependent rule that remains useful for any horizon the handoff policy may pick, and the shared weights let it run for horizons outside $\mathcal{H}$. We initialize the output layer of $f_\theta$ at zero, so that $(u_k^{\mathrm p}, u_k^{\mathrm d}) = (1,1)$ and training starts from the solver's native step-size rule.

\textbf{Training the handoff policy.} Unlike the distance used to train the predictor, the end-to-end time is measured in wall-clock and is not differentiable. We therefore freeze the trained predictor, roll it out on each training instance, and measure at every warm-start iteration $k$ the end-to-end time of handing off at $k$, divided by the cold-start time, $
  C_i(k) = \mathcal{T}_i(\theta, k; \epsilon) \,/\, \mathcal{T}_{i,\mathrm{cold}},$ using actual solver runs from the warm start rather than a surrogate. The regression targets for the two predictions of the handoff policy are
\begin{equation}
  Q_{i,\mathrm{handoff}}(k) = C_i(k),
  \qquad
  Q_{i,\mathrm{continue}}(k) = \min_{k<j\le K_{\max}} C_i(j),
  \label{eq:q-targets}
\end{equation}
the normalized end-to-end time of handing off now and the best one attainable by
handing off later. We fit $g_\varphi$ by
\begin{equation}
  \min_\varphi\;
  \frac{1}{|\mathcal{D}_{\mathrm{handoff}}|}
  \sum_{i\in\mathcal{D}_{\mathrm{handoff}}}
  \frac{1}{|\mathcal{J}_i|}
  \sum_{k\in\mathcal{J}_i}
  \sum_{a\in\{\mathrm{handoff},\,\mathrm{continue}\}}
  \big(\widehat{Q}_{a} (\mathbf{h}_{i,k}) - Q_{i,a}(k)\big)^2,
  \label{eq:q-loss}
\end{equation}
where $\mathcal{D}_{\mathrm{handoff}}$ is the set of training instances for the handoff policy and $\mathcal{J}_i$ is the set of warm-start iterations evaluated on instance $i$. Because both outputs are regressed onto the same normalized end-to-end time, the handoff rule compares them directly, without a hand-tuned threshold.
\section{Experiments}
\label{sec:experiments}

In this section, we first examine the proposed adaptive accelerator against common baselines, and further examine the generalization capabilities, adaptive stopping and difficulty. Finally, we conduct ablation analysis to analyze the underlying structure of AHEAD.

\subsection{Experimental Settings}
\label{subsec}

\textbf{Datasets and metrics.} We evaluate on two LP families: synthetic packing LPs with $n\in[10^5,10^6]$ nonnegative variables, one equality constraint, and about $0.1n$ sparse inequalities, and PageRank LPs on Barabási--Albert graphs~\citep{applegate2021practical,iouditski2014primaldualsubgradientmethodsminimizing}. Both follow the LP form in Section~\ref{sec:cost}. We report the end-to-end speedup, whose wall-clock time includes AHEAD's inference, presolve, and solver execution, and the reduction in iterations, both at the same target accuracy ($\epsilon=10^{-4}$ by default). Test instances are disjoint from the training instances. Appendices~\ref{app:pagerank}, \ref{sec:synthetic_packing}, and~\ref{app:exp} give the generation details and the measurement protocol.

\textbf{Baselines and Evaluation.} We compare against three groups of baselines. Cold-start solvers form the first group: Gurobi GPU PDHG (\texttt{PDHGGPU=1}) and OR-Tools PDLP. The second group is the learned baseline PDHG-Net~\citep{li2024pdhg}, the closest learned method to ours. The third group contains two variants of AHEAD: a fixed horizon selected on validation, and a per-instance optimal $K_i^\star$ that marks the best achievable handoff.

\textbf{Training Details.} The candidate horizons are $\mathcal{H} = \{8, 16, 32, 64, 128\}$. We train the predictor for 200 epochs with Adam, a learning rate of $10^{-3}$, cosine annealing, and a batch size of $8$. Each mini-batch shares one sampled horizon $K$, which enables batched rollouts and still gives an unbiased estimate of Equation~\ref{eq:multi-horizon}. We clip gradients so that pathological primal or dual errors do not dominate an update. The handoff policy is then trained on the frozen predictor's trajectories.

\subsection{Acceleration}
Table~\ref{tab:accel-summary} compares AHEAD with PDHG-Net and with a fixed number of warm-start iterations selected on validation, all on the Gurobi GPU-PDHG backbone at $\epsilon=10^{-4}$.
\begingroup
\setlength{\columnsep}{5pt}
\setlength{\intextsep}{3pt}
\needspace{14\baselineskip}
\begin{wrapfigure}[13]{r}{0.38\textwidth}
    \vspace{1\baselineskip}
    \centering
    \includegraphics[width=\linewidth]{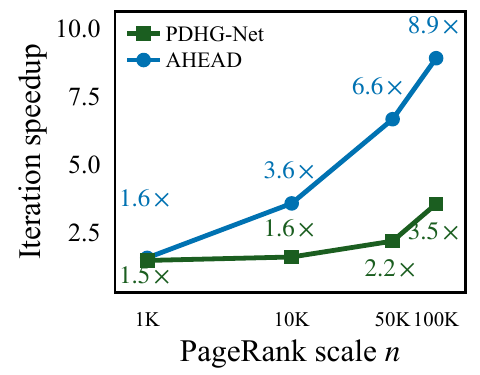}
    \captionsetup{skip=5pt}
    \caption{Speedup on PageRank LPs.}
    \label{fig:pagerank}
\end{wrapfigure}
The test sets contain 50 held-out packing LPs, 20 larger packing LPs with $n \in [10^6,10^7]$, and 20 packing LPs from an unseen distribution. (1) AHEAD generalizes across scale and distribution. Its predictor only rescales step sizes from a fixed-dimensional solver state and never predicts the solution, unlike PDHG-Net, whose predictions are tied to the training distribution. (2) Adaptive handoff nearly matches the best handoff. The best number of warm-start iterations varies across instances, so a fixed number wastes iterations or stops too early, whereas the handoff policy stops once further iterations no longer save time. (3) The advantage grows with problem size. Larger PageRank LPs require more iterations, and the weight-shared predictor adapts to each instance, whereas PDHG-Net has fixed depth (Figure~\ref{fig:pagerank}).
\par
\endgroup

\begin{table*}[!ht]
\vspace{6pt}
\centering
\caption{Acceleration with the Gurobi PDHG backbone on packing-type LPs.}
\label{tab:accel-summary}
\vspace{-8pt}

\setlength{\tabcolsep}{4pt}
\renewcommand{\arraystretch}{1.05}
\footnotesize

\begin{tabular*}{\textwidth}{@{\extracolsep{\fill}}lcccccc@{}}
\toprule
\multirow{2}{*}[-2pt]{\textbf{Method}}
& \multicolumn{2}{c}{\textbf{IID} ($n{=}50$)}
& \multicolumn{2}{c}{\textbf{Size shift} ($n{=}20$)}
& \multicolumn{2}{c}{\textbf{Unseen distribution} ($n{=}20$)} \\
\cmidrule(lr){2-3}
\cmidrule(lr){4-5}
\cmidrule(lr){6-7}
& Speedup & Time (s)
& Speedup & Time (s)
& Speedup & Time (s) \\
\midrule
Cold PDHG (Gurobi)
& $1.00\times$ & $18.2$
& $1.00\times$ & $5.3$
& $1.00\times$ & $3.2$ \\

Best validation $K$ ($K{=}32$)
& $5.63\times$ & $3.2$
& $5.09\times$ & $1.1$
& $3.25\times$ & $0.99$ \\

PDHG-Net~\citep{li2024pdhg}
& $1.23\times$ & $14.8$
& $1.18\times$ & $4.5$
& $1.06\times$ & $3.1$ \\
\midrule
\textbf{AHEAD}
& $\mathbf{11.07\times}$ & $\mathbf{1.6}$
& $\mathbf{11.59\times}$ & $\mathbf{0.46}$
& $\mathbf{6.48\times}$ & $\mathbf{0.50}$ \\

\textit{Per-instance optimal $K^\star$}
& $12.36\times$ & $1.5$
& $12.75\times$ & $0.42$
& $7.06\times$ & $0.46$ \\
\bottomrule
\end{tabular*}

\vspace{0.5pt}
\end{table*}

\subsection{Generalization}
\begin{figure}[!ht]
    \centering
    \captionsetup[subfigure]{skip=5pt}

    \begin{subfigure}[t]{0.32\linewidth}
        \centering
        \includegraphics[
            width=\linewidth,
            trim=0 0pt 0 0,
            clip
        ]{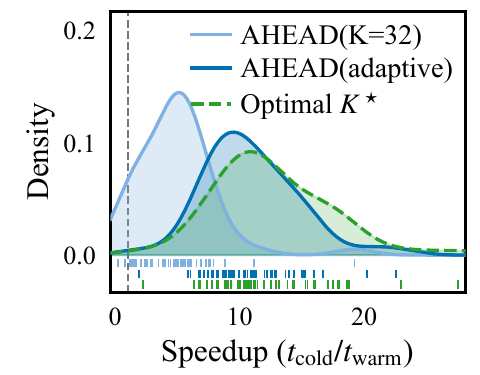}
        \caption{Fixed vs.\ adaptive handoff.}
        \label{fig:fixed_vs_adaptive_k}
    \end{subfigure}
    \hfill
    \begin{subfigure}[t]{0.32\linewidth}
        \centering
        \includegraphics[
            width=\linewidth,
            trim=0 0pt 0 0,
            clip
        ]{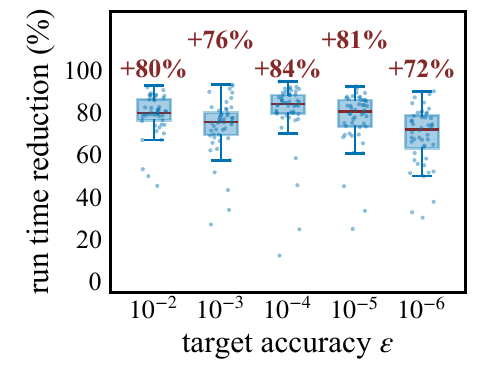}
        \caption{Robustness across target acc.}
        \label{fig:eps_sweep}
    \end{subfigure}
    \hfill
    \begin{subfigure}[t]{0.32\linewidth}
        \centering
        \includegraphics[
            width=\linewidth,
            trim=0 0pt 0 0,
            clip
        ]{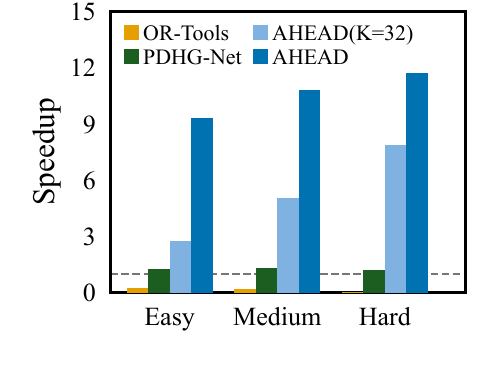}
        \caption{Speedup across difficulty tiers.}
        \label{fig:difficulty}
    \end{subfigure}

    \captionsetup{skip=5pt}
    \caption{
        Robustness across acceleration horizons, target accuracies,
        and problem difficulty levels.
    }
    \label{fig:robustness_and_adaptation}
\end{figure}

\begingroup
\setlength{\columnsep}{5pt}
\setlength{\intextsep}{3pt}
\begin{wrapfigure}{r}{0.38\textwidth}
    \centering
    \includegraphics[width=\linewidth]{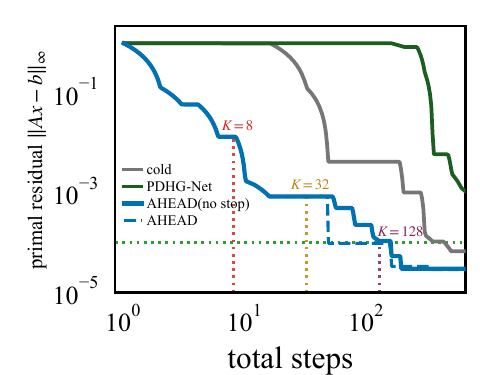}
    \captionsetup{skip=5pt}
    \caption{Primal-residual trajectories without early solver handoff.}
    \label{fig:precision_curve}
\end{wrapfigure}
\textbf{Does AHEAD rely on solver continuation?} We disable handoff on a representative LP instance and compare AHEAD's residual trajectory with adaptive AHEAD, cold-start PDHG, and PDHG-Net. Figure~\ref{fig:precision_curve} supports two observations. (1) AHEAD can reach the target tolerance without native continuation. When AHEAD runs without handoff, it reduces the primal residual and reaches the target, whereas PDHG-Net remains above it. (2) Learned acceleration and solver refinement are complementary. Adaptive handoff reaches the target earlier than AHEAD without handoff or the cold-start solver, indicating that AHEAD accelerates early trajectory while the native solver completes the refinement. Handoff improves efficiency rather than providing progress that AHEAD cannot achieve on its own.
\par
\endgroup

\textbf{Adaptive Stopping.} To evaluate whether AHEAD can select an effective number of warm-start iterations, we compare adaptive handoff with a fixed $K=32$ and the per-instance optimal $K^\star$, the $K\in\{1,\ldots,128\}$ with the shortest end-to-end time on each instance; we also evaluate robustness across target accuracies. The figures support two observations. (1) Adaptive handoff comes close to the per-instance optimal $K^\star$. In Figure~\ref{fig:fixed_vs_adaptive_k}, the adaptive speedup distribution shifts markedly beyond the fixed-horizon baseline and toward the oracle, showing that the current solver state provides useful information for selecting an instance-specific handoff point. (2) Acceleration persists across target accuracies. Figure~\ref{fig:eps_sweep} shows consistently large runtime reductions across all evaluated tolerances. Because every run must satisfy the same PDLP termination criterion, these gains demonstrate consistent speedups for convergence without relaxing final solution accuracy.

\textbf{Difficulty.} To test whether acceleration persists as instances become harder, we generate 60 packing-LP test instances unseen during training and divide them into three equal-sized groups, Easy, Medium, and Hard, based on their cold-start Gurobi PDHG runtimes. We then split them into three equal-sized groups: the shortest third as Easy, the middle third as Medium, and the longest third as Hard. Figure~\ref{fig:difficulty} reports \textsc{AHEAD} achieves approximately $9\times$, $11\times$, and $12\times$ speedup on easy, medium, and hard instances, respectively. The corresponding fixed $K=32$ results are roughly $3\times$, $5\times$, and $8\times$. Adaptive handoff improves speedup in every tier, while OR-Tools PDLP and PDHG-Net remain below the cold Gurobi baseline in this comparison.

\subsection{Ablation}
\begin{wraptable}{r}{0.50\textwidth}
    \vspace{-0.5\baselineskip}
    \centering
    \caption{
        Incremental ablation of solver-state features. Features are added cumulatively.
    }
    \label{tab:ablation_features}
    \vspace{-8pt}

    \setlength{\tabcolsep}{4pt}
    \renewcommand{\arraystretch}{1.05}
    \scriptsize

    \begin{tabular*}{\linewidth}{
        @{\extracolsep{\fill}}lcc@{}
    }
        \toprule
        \textbf{Input state}
        & \makecell{\textbf{Total} \textbf{steps} $\downarrow$}
        & $\boldsymbol{\Delta}$ (\%) $\uparrow$ \\
        \midrule
        Fixed stepsize
        & 104.4 & -- \\

        $+$ Residuals
        & 92.5 & 11.4 \\

        $+$ Movement/nonlinearity
        & 80.4 & \textbf{13.1} \\

        $+$ Primal weight $\omega_k$
        & 79.3 & 1.4 \\

        $+$ Cosine geometry (AHEAD)
        & \textbf{76.8} & 3.2 \\
        \bottomrule
    \end{tabular*}

    \vspace{0.5pt}
\end{wraptable}

\textbf{Which solver-state features matter?} Table~\ref{tab:ablation_features} incrementally introduces the four groups of solver-state features. Adding the primal and dual residuals reduces the total number of steps from $104.4$ to $92.5$. Movement and nonlinearity provide the largest incremental improvement, further reducing the total to $80.4$ steps. Incorporating the primal weight and cosine features subsequently reduces the total to $79.3$ and $76.8$, respectively. Overall, the complete representation achieves a $26.4\%$ reduction relative to the fixed step-size baseline. These results indicate that feasibility, local stability, primal-dual scaling, and trajectory geometry provide complementary information for step-size prediction.

\par\medskip

\begin{wraptable}{r}{0.50\textwidth}
    \vspace{-0.5\baselineskip}
    \centering
    \caption{
        Predictor-architecture ablation.
    }
    \label{tab:ablation_architecture}
    \vspace{-8pt}

    \setlength{\tabcolsep}{4pt}
    \renewcommand{\arraystretch}{1.05}
    \scriptsize

    \begin{tabular*}{\linewidth}{
        @{\extracolsep{\fill}}lcc@{}
    }
        \toprule
        \textbf{Controller}
        & \makecell{\textbf{Total} \textbf{steps} $\downarrow$}
        & \makecell{\textbf{Continuation} \textbf{steps} $\downarrow$} \\
        \midrule
        5 layers, width 64
        & 68.5 & 36.5 \\

        7 layers, width 128
        & 72.9 & 40.9 \\

        Weight-shared (AHEAD)
        & \textbf{67.8} & \textbf{35.8} \\
        \bottomrule
    \end{tabular*}

    \vspace{0.5pt}
\end{wraptable}

\textbf{Does the improvement come from model capacity?} Table~\ref{tab:ablation_architecture} compares controller architectures while fixing the solver-state representation and training objective. Increasing the controller from five layers with width $64$ to seven layers with width $128$ increases the total number of steps from $68.5$ to $72.9$. The improvement therefore cannot be explained simply by increased network capacity. In contrast, sharing controller parameters across iterations produces the best result, requiring $67.8$ total steps and $35.8$ continuation steps. This supports matching the controller parameterization to the iteration-invariant update structure of the underlying optimizer.

\par\medskip

\section{Conclusion}
We asked what neural accelerator design achieves end-to-end speedups on large-scale LPs across sizes and target accuracies. AHEAD answers with three design choices. First, it learns only the step sizes and keeps the solver's update directions, so each warm-start iteration stays inexpensive and every iterate remains a valid warm start. Second, its weight-shared predictor runs recurrently, so the warm-start phase can last any number of iterations. Third, a learned handoff policy selects this number for each instance and accuracy. Together, these choices let the solver reach the target accuracy in fewer iterations with little time per iteration, yielding many-fold end-to-end speedups on packing and PageRank LPs across unseen scales, difficulty levels, and target accuracies. Future work will extend AHEAD to real-world LPs and other first-order solvers.

\clearpage
\bibliography{iclr2027_conference}
\bibliographystyle{iclr2027_conference}

\newpage
\appendix
\section{Engineering Accelerations in PDLP}
\label{app:pdlp}

Classical PDHG uses fixed primal and dual step sizes chosen according to a global stability condition, making its convergence highly sensitive to conservative step-size selection and mismatched primal–dual scaling. 
Beyond presolving and diagonal preconditioning, PDLP accelerates conventional PDHG through three coupled mechanisms: \textbf{adaptive step-size selection}, \textbf{adaptive restarting}, and \textbf{dynamic primal weighting}. 

First, PDHG parameterizes its primal and dual step sizes as$$
\tau_k=\frac{\eta_k}{\omega_k}, \quad \sigma_k=\eta_k \omega_k
$$

where $\eta_k$ controls the overall step magnitude and $\omega_k$ controls the relative primal-dual scaling. Conventional PDHG selects $\eta_k$ according to the global condition $\eta_k \leq 1 /\|A\|_2$, which guarantees convergence but is often overly conservative. PDLP instead evaluates each candidate step using the local condition
$
\eta_k \leq \bar{\eta}_k:=\frac{\left\|z_{k+1}-z_k\right\|_{\omega_k}^2}{2\left(y_{k+1}-y_k\right)^{\top} A\left(x_{k+1}-x_k\right)},
$ where $z_k=\left(x_k, y_k\right)$ and $
\|z\|_\omega^2=\omega\|x\|_2^2+\frac{1}{\omega}\|y\|_2^2 .$ The next candidate step size is updated as $
\eta_{k+1}=\min \left\{\left(1-(k+1)^{-0.3}\right) \bar{\eta}_k,\left(1+(k+1)^{-0.6}\right) \eta_k\right\} .$ A candidate violating the local condition is rejected and recomputed using a smaller step. Because $\bar{\eta}_k$ depends on the realized primal-dual movement rather than only on the global operator norm, PDLP can take substantially larger steps along locally stable portions of the trajectory.

The second mechanism \textbf{controls trajectory history} through adaptive restarting. Within outer iteration $n$, PDLP maintains a step-size-weighted average of the inner iterates, $
\bar{z}^{n, t+1}=\frac{\sum_{i=1}^{t+1} \eta^{n, i} z^{n, i}}{\sum_{i=1}^{t+1} \eta^{n, i}} .$ Progress is evaluated using the normalized duality gap 
$$
\rho_r^n(z)=\frac{1}{r} \max _{\substack{\hat{z}=(\hat{x}, \hat{y}) \in \mathcal{Z} \\\|\hat{z}-z\|_{\omega^n} \leq r}}\{\mathcal{L}(x, \hat{y})-\mathcal{L}(\hat{x}, y)\},$$
and $
\mu_n\left(z, z_{\text {ref }}\right)=\rho_{\left\|z-z_{\text {ref }}\right\|_{\omega^n}}(z) .$

At each restart check, PDLP selects between the current iterate and its weighted average:

$$
z_c^{n, t+1}= \begin{cases}z^{n, t+1}, & \mu_n\left(z^{n, t+1}, z^{n, 0}\right)<\mu_n\left(\bar{z}^{n, t+1}, z^{n, 0}\right) \\ \bar{z}^{n, t+1}, & \text { otherwise }\end{cases}
$$

The outer loop is restarted when the normalized duality gap exhibits sufficient decay, when local progress stalls after necessary decay has been achieved, or when the inner loop becomes excessively long. This mechanism retains averaging when it is beneficial but discards accumulated trajectory history once it ceases to improve the normalized gap.

Finally, PDLP updates the \textbf{primal weight} only at restart points. Let $
\Delta_x^n=\left\|x^{n, 0}-x^{n-1,0}\right\|_2, \quad \Delta_y^n=\left\|y^{n, 0}-y^{n-1,0}\right\|_2 .$ The new primal weight is obtained through log-space smoothing: $
\omega^n=\exp \left(\theta \log \frac{\Delta_y^n}{\Delta_x^n}+(1-\theta) \log \omega^{n-1}\right),$ where PDLP uses $\theta=0.5$. This update approximately balances the weighted primal and dual distances while avoiding abrupt changes that may destabilize PDHG. Consequently, the three mechanisms address distinct limitations of conventional PDHG: adaptive step sizes control the step magnitude using local stability, restarts manage accumulated trajectory history, and the primal weight balances progress between the primal and dual spaces.

\section{Training}
\begin{algorithm}[H]
\caption{Random-Horizon Training}
\label{alg:kfree-ddpm}
\begin{algorithmic}[1]
\Require Dataset $\mathcal{D}$ with targets
$(\mathbf{x}^{\star},\mathbf{y}^{\star})$;
$\mathcal{K}=\{8,16,32,64,128\}$; batch size $B$
\State \textbf{Initialization} 
\Repeat
    \State Sample a mini-batch $\mathcal{B}\subset\mathcal{D}$ and a shared
    $K\sim\operatorname{Uniform}(\mathcal{K})$
    \ForAll{$i\in\mathcal{B}$}
        \State $(\hat{\mathbf{x}}_{i}^{K},\hat{\mathbf{y}}_{i}^{K})
        \gets \operatorname{Rollout}_{K}(f_{\theta},i)$
        \State $d_i \gets
        \dfrac{\|\hat{\mathbf{x}}_{i}^{K}-\mathbf{x}_{i}^{\star}\|_2^2}
              {\|\mathbf{x}_{i}^{\star}\|_2^2+\delta}
        +
        \dfrac{\|\hat{\mathbf{y}}_{i}^{K}-\mathbf{y}_{i}^{\star}\|_2^2}
              {\|\mathbf{y}_{i}^{\star}\|_2^2+\delta}$
    \EndFor
    \State $\mathcal{L}_{10^{-8}}
    \gets \dfrac{1}{B}\sum_{i\in\mathcal{B}}
    d_i$
    \State Update $\theta$ using $\nabla_{\theta}\mathcal{L}_{10^{-8}}$
\Until{convergence}
\end{algorithmic}
\end{algorithm}

\section{Dataset}
\subsection{PageRank}
\label{app:pagerank}
Following \citet{applegate2021practical}, we generate
PageRank LPs with the formulation of
\citet{iouditski2014primaldualsubgradientmethodsminimizing}, using a damping
factor $\lambda = 0.85$. The resulting LP has one equality row
$\mathbf{1}^\top x = 1$ and $n$ inequality rows, and matches the LP form in
Section~\ref{sec:cost} with $S = 1$ and the transition constraints written as
sparse linear inequalities.

For each instance, we first sample a Barabási--Albert
preferential-attachment graph with attachment degree $3$.
Let $S'\in\mathbb{R}^{n\times n}$ denote its column-normalized
adjacency matrix, such that $S'$ is column stochastic.
Given a damping factor $\lambda\in(0,1)$, the PageRank transition
matrix is
\begin{equation}
    S
    =
    \lambda S'
    +
    \frac{1-\lambda}{n}
    \mathbf{1}\mathbf{1}^{\top}.
    \label{eq:pagerank_transition}
\end{equation}
The corresponding PageRank vector can be obtained as a feasible
solution to
\begin{equation}
\begin{aligned}
    \mathrm{find}\quad
        & x,                                                    \\
    \mathrm{subject~to}\quad
        & Sx \leq x,                                            \\
        & \mathbf{1}^{\top}x = 1,                               \\
        & x \geq 0.
\end{aligned}
\label{eq:pagerank_lp}
\end{equation}
Directly constructing $S$ would introduce a dense teleportation
matrix. Using the normalization constraint
$\mathbf{1}^{\top}x=1$, we instead retain the equivalent sparse
constraints
\begin{equation}
    \lambda(S'x)_i
    +
    \frac{1-\lambda}{n}
    \leq x_i,
    \qquad
    i=1,\ldots,n.
    \label{eq:pagerank_sparse}
\end{equation}

\subsection{Synthetic Packing LPs}
\label{sec:synthetic_packing}

We construct a family of synthetic packing LPs of the form
\begin{equation}
\label{eq:synthetic_packing_lp}
\begin{aligned}
    \min_{x \in \mathbb{R}^{n}} \quad
        & c^{\top}x \\
    \text{s.t.} \quad
        & \mathbf{1}^{\top}x = S, \\
        & Ax \leq b, \\
        & x \geq 0 .
\end{aligned}
\end{equation}
Each instance has $n \in [10^5, 10^6]$ variables and $m = 0.1n$ inequality
constraints. The constraint matrix is generated as
\begin{equation}
    A_{ij}=M_{ij}\widetilde{A}_{ij},
    \qquad
    M_{ij}\sim\operatorname{Bernoulli}(\rho),
    \qquad
    \widetilde{A}_{ij}\sim\operatorname{Unif}(0.5,1.0),
\end{equation}
with density $\rho = 100/n$, so each inequality involves about $100$ variables
on average. The variables have no explicit upper bounds, and the instances match
the LP form in Section~\ref{sec:cost} directly. The objective coefficients are
sampled as $c_j \sim \operatorname{Unif}(-1, 2)$.

For each instance, we first sample the equality right-hand side as
$S\sim\operatorname{Unif}(0.3,0.7)$. We then draw
\begin{equation}
    u\sim\operatorname{Dirichlet}(\mathbf{1}_{n}),
    \qquad
    x_{0}=Su,
\end{equation}
such that $x_{0}\geq 0$ and $\mathbf{1}^{\top}x_{0}=S$. The inequality
right-hand side is constructed as
\begin{equation}
    b=Ax_{0}+s,
\end{equation}
where the entries of $s$ are sampled independently. This construction makes
$x_{0}$ feasible and controls instance difficulty through the magnitude of the
available slack: larger slack values produce mostly inactive inequalities,
whereas smaller values yield a larger fraction of binding constraints at the
optimum. 

\paragraph{Data splits.}
The training instances and the in-distribution test set are drawn from the
generator above with $n \in [10^5, 10^6]$. The size-shift test set contains 20
instances with $n \in [10^6, 10^7]$ from the same generator. The
distribution-shift test set contains 20 instances whose generation parameters
differ from those used for training. The difficulty tiers in
Figure~\ref{fig:difficulty} group 60 held-out instances by their cold-start
time. Training, validation, and test instances are generated using disjoint
random seeds and checked for duplicate content. For the difficulty study (Figure~\ref{fig:difficulty}), we sort all test instances by their cold-start time on Gurobi PDHG and split them into three equal-sized groups: the shortest third as Easy, the middle third as Medium, and the longest third as Hard.

\subsection{Experimental Environment and Measurement Protocol}
\label{app:exp}

All experiments were conducted on the same machine.
Table~\ref{tab:environment} summarizes the hardware and software environment.

\begin{table}[!th]
\centering
\caption{Experimental environment.}
\label{tab:environment}
\small
\setlength{\tabcolsep}{5pt}
\begin{tabular}{@{}p{0.22\linewidth}p{0.68\linewidth}@{}}
\toprule
\textbf{Component} & \textbf{Configuration} \\
\midrule
GPU & NVIDIA A100 PCIe 40GB; driver 580.173.02 \\
CPU & 2$\times$ Intel Xeon Gold 5317; 24 cores, 48 threads \\
Memory & 251 GiB \\
OS & Ubuntu 24.04.3 LTS; Linux 6.8.0-139 \\
Python & 3.12.9 \\
PyTorch & 2.9.0 \\
CUDA toolkit & 12.0 (V12.0.140) \\
OR-Tools & 9.15.6755 \\
Gurobi & 13.0.3, GPU-enabled build \\
\bottomrule
\end{tabular}
\end{table}

\paragraph{Timing metric.}
Unless stated otherwise, the aggregate end-to-end speedup is
\begin{equation}
\mathrm{Speedup}
=
\frac{\sum_i \mathcal{T}_{i,\mathrm{cold}}}
     {\sum_i \mathcal{T}_i(\theta, K_i; \epsilon)},
\end{equation}
where $\mathcal{T}_i = T_{\mathrm{warm}} + T_{\mathrm{solver}}$ is the
end-to-end time defined in Section~\ref{sec:method}.

\paragraph{Warm-start iterations and handoff.}
The number of warm-start iterations $K$ counts complete PDHG updates whose
step-size multipliers are predicted by AHEAD. After $K$ iterations, AHEAD passes
the current primal iterate $\hat{\mathbf{x}}_K$ and dual iterate
$\hat{\mathbf{y}}_K$ to the solver as its initial primal and dual solution; for
Gurobi, these are assigned through \texttt{PStart} and \texttt{DStart}. No
predictor state or step size is transferred, and the solver continues with its
own update rule. The warm-start iterations are included in the end-to-end time.
The fixed number of warm-start iterations used as a baseline is selected on
validation data.

\end{document}